\documentclass[conference]{IEEEtran}
\IEEEoverridecommandlockouts
\usepackage{cite}
\usepackage{amsmath,amssymb,amsfonts}
\usepackage{algorithmic}
\usepackage{graphicx}
\usepackage{textcomp}
\usepackage{xcolor}
\usepackage{listings}
\usepackage{eurosym}
\usepackage{subcaption}
\def\BibTeX{{\rm B\kern-.05em{\sc i\kern-.025em b}\kern-.08em
    T\kern-.1667em\lower.7ex\hbox{E}\kern-.125emX}}
\begin{document}

\title{Ludii as a Competition Platform\\
}

\author{\IEEEauthorblockN{Matthew Stephenson, {\'E}ric Piette, Dennis J.N.J. Soemers and Cameron Browne}
\IEEEauthorblockA{\textit{Department of Data Science and Knowledge Engineering} \\
\textit{Maastricht University}\\
Maastricht, The Netherlands \\
\texttt{\{matthew.stephenson,eric.piette,dennis.soemers,cameron.browne\}@maastrichtuniversity.nl}}
}

\maketitle

\begin{abstract}
Ludii is a general game system being developed as part of the ERC-funded Digital Ludeme Project (DLP). While its primary aim is to model, play, and analyse the full range of traditional strategy games, Ludii also has the potential to support a wide range of AI research topics and competitions. This paper describes some of the future competitions and challenges that we intend to run using the Ludii system, highlighting some of its most important aspects that can potentially lead to many algorithm improvements and new avenues of research. We compare and contrast our proposed competition motivations, goals and frameworks against those of existing general game playing competitions, addressing the strengths and weaknesses of each platform.


\end{abstract}

\begin{IEEEkeywords}
General Game Playing, Artificial Intelligence, Competitions, Ludii, Ludemes, Board games
\end{IEEEkeywords}

\section{Introduction}

Game-based research and analysis has, and continues to be, one of the most popular and long running approaches for developing, evaluating and comparing new artificial intelligence algorithms \cite{gameaibook}. Since the early days of AI research, games have been used to demonstrate the abilities and potential of AI for solving complex combinatorial problems \cite{turing}. Playing games effectively often requires multiple skills, such as understanding the game rules, evaluating different situations, planning several moves ahead, predicting the behaviour of other players, selecting one of many possible actions, and combining all these aspects together to determine an optimal strategy.

\subsection{Game Playing Agents}

Many algorithms have been developed for playing specific games with the goal of being able to outperform human experts, achieving super-human performance. This includes many board or card games such as Chess \cite{stockfish17}, Go \cite{silver16}, Shogi \cite{shogiSilver}, Checkers \cite{checkers}, Backgammon \cite{tdgammon}, Hanabi \cite{hanabi} and Poker \cite{pokerreview}, as well as popular video games such as Super Mario Bros. \cite{marioaicompetition}, Starcraft \cite{starcraft}, Angry Birds \cite{aibirdscompetition}, Ms. PacMan \cite{pacman}, Doom \cite{vizdoom} and Unreal Tournament \cite{unreal}. Many of these games are also associated with competitions that hope to provide a benchmark comparison between the best approaches currently being developed \cite{gamecompetitions}. Such competitions are a great way to encourage collaborations and friendly rivalries between different teams and AI techniques. 

Apart from being highly enjoyable to watch, developing agents that can play traditional or modern games effectively has many immediate benefits and real-world applications \cite{gameaibook}. 
Research on games has led to new and improved AI techniques that are applicable across multiple scenarios. Examples include the many recent advances in Deep Learning and Monte Carlo Tree Search (MCTS) promoted by agents such as AlphaGo and AlphaGo Zero for the game of Go \cite{alphago,alphagozero}, the use of counterfactual regret minimization and subgame abstraction for playing both limit and no-limit Texas hold'em poker respectively \cite{pokerlimit,pokernolimit}, and the use of parallelized Alpha-Beta pruning for playing Chess \cite{chessalphabeta}. Such research can also be applied to many non-game related problems \cite{gamesfriends}, validating the use of games as a well defined and safe benchmark task for evaluating new AI techniques.

\subsection{General Game AI}
While many of these game specific agents and techniques can be very effective for their intended domain, developing agents that are only capable of playing a specific or limited subset of games has its downsides. Many of these game playing algorithms tend to over-specialise on the game they are designed for, as domain specific knowledge and highly tailored heuristics can often give a substantial performance advantage \cite{domain,hearthsonedomain}. Because of this, such agents are typically unable to transfer their abilities between different games. A world class poker agent is unlikely to be able to play Chess or Go with anywhere near the same level of skill that it previously exhibited. This is a significant limitation of current AI agents compared to most human players, as many of us are likely able to play a wide range of games and can learn new ones relatively quickly. The field of general game playing attempts to address this problem by requiring developed agents to work well across many different, and sometimes unknown, game possibilities. This has become one of the grand challenges of AI: Creating general intelligence that is capable of solving many different tasks, or in the case of general game playing, capable of playing many different games. 

Apart from simply playing a given game as well as possible, many alternative AI problems focused around general games also exist.

\subsubsection{Human-like AI}

 Instead of optimal performance, some researchers try to develop agents that achieve human-like game playing abilities \cite{humanai}. These agents try to mimic the playing style of human players, who often are not as adept as AI in some game aspects, such as checking many possible scenarios or performing complex mathematical calculations, but are likely to be better or equal in other areas. Such agents typically require fewer computational resources than those needed to achieve super-human performance, and would likely be much better at evaluating how enjoyable or well-balanced a particular game might be for a human player \cite{personalisedcontent}.

\subsubsection{Procedural Content Generation}

Another popular area of general game research is that of procedural content generation (PCG). Most PCG algorithms are developed specifically for generating a particular game element (level, weapon, vehicle, objective, etc.) for a specific type of game \cite{pcg1,pcg2,pcg3}. General PCG takes these same concepts but applies them to a much wider range of applicable games types. This can also include generating entirely new games from scratch, or generating certain game aspects given other pre-set requirements. Developing level or rule generators that can successfully create well-designed and enjoyable content for a wide range of different games creates many new difficulties beyond that of standard PCG \cite{generallevels,generalrules}. General game evaluation metrics should also be able to assess the quality of multiple games for a variety of different factors, such as enjoyment, content variety, strategic depth, expected playtime, etc.

\subsection{General Game Platforms}

Several research platforms currently exist for promoting the development of general AI for games \cite{swiechowski15,generalgame2}. The most notable examples of this include the General Game Playing (GGP) system \cite{genesereth05}, which focuses primarily on board and abstract games, the General Video Game AI (GVGAI) framework \cite{gvgaioverview}, which focuses on simple video games, and the Arcade Learning Environment \cite{aleoverview,aleoverview2}, which allows agents to play a large number of Atari 2600 games. Both the GGP and GVGAI platforms also have competitions associated with them that promote and compare work done using their respective systems, and which are discussed in greater detail in the following section.

Ludii is an upcoming ludemic general game system that hopes to provide a large number of exciting opportunities for general game AI research \cite{ludii1}. Ludii has so far been shown to possess many improvements in efficiency, simplicity and generality beyond those of other prior general game systems, and the ludemic approach it takes for describing games allows for many new research possibilities. We describe in this paper some of the competitions that we intend to organise and run using it. While some of these described competitions overlap the structures and goals of existing alternatives, we believe that the benefits of our system, combined with the novel and varied ideas we present, will make Ludii one of the most attractive and popular competition platforms in future years.

\subsection{Paper Overview}

The remainder of this paper is organised as follows: Section II describes some existing general game competitions, the different tracks they offer, as well as their respective strengths and weaknesses. Section III describes our new proposed Ludii system in greater detail, highlighting some of its most important and appealing aspects that make it suitable for a wide range of competitions. Section IV describes some of the agent-based competitions that we intend to run with Ludii. Section V describes some alternative PCG-based competitions that can either run separately or alongside the agent-based competitions. Section VI provides some possible competitions which do not strictly feature agents or PCG, instead delving into other areas of AI research such as data mining, player analysis, and information visualisation. Section VII finishes by concluding and summarising the ideas presented within this paper, as well as addressing any potential challenges or shortcomings that could occur.

\section{Existing General Game Competitions}
The main competitions around developing general game AI are the International General Game Playing (IGGP) Competition and the General Video game AI (GVGAI) Competition. While some other general game competitions exist, such as the Tiltyard Open competition,\footnote{http://tiltyard.ggp.org/} these two are easily the most popular and well-documented examples. In this section we discuss both of these competitions, describing the types of games they focus on, how the competitions are structured, and the different tracks they offer.

\subsection{IGGP Competition}
The IGGP competition was one of the first general game playing competitions, organised
by the Logic group of the Stanford University in 2005 \cite{genesereth13}. This competition is hosted annually at the AAAI conference, and centers around the creation of agents that can play games defined using the Game Description Language (GDL) \cite{love08}. Games written in GDL are described in terms of simple instructions based on first-order logic clauses. This restricts the types of games that can be written in the base GDL language to deterministic games with perfect information. While extensions to this language have been proposed that allow for imperfect information and epistemic games \cite{schiffel14,thielscher17}, these are currently not included as part of the base GDL game repository\footnote{GGPBASE: https://github.com/ggp-org/ggp-base} and are not used in the IGGP competition. This repository is typically updated with a few extra games each year.

Compared to other alternative languages, GDL provides a low-level description of games. Even the simplest of games can take hundreds or thousands of lines to define, and making small modifications to existing games, such as changing the size or shape of the board, often requires many lines of code to be changed or added. Presenting only the underlying logic for the game, rather than encapsulating common game concepts within more human-understandable terms, makes game descriptions difficult for humans to understand. Processing these descriptions can also be computationally expensive due to the logic resolution required, and GDL agents must construct their own internal representation of each game's mechanics in order to play effectively. Many medium to high complexity games either require a very large amount of time to model (e.g. Go) or are rendered unplayable due to the large computational costs involved (e.g. Chess). Even more complex board games such as Arimaa \cite{arimaa}, which was designed specifically to be difficult for brute force tree search algorithms, would be almost impossible to describe in GDL, likely requiring thousands of lines of code.

The IGGP competition has a single track, that focuses exclusively on playing games. 
Agents submitted to the competition are evaluated on a pre-determined collection of several previously unseen GDL games. Most of the games written in GDL, as well as the majority of games used in the competition, tend to be variants of well known board or puzzle games. Competition entrants do not know which types of games their agent will be required to play beforehand. During the competition, each participating agent initially receives the GDL description of each game it is required to play, often between 30 and 120 seconds beforehand, and is typically given between 5 and 30 seconds of decision time for each move that it makes. The competition consists of an initial preliminary round that is open to everyone, after which the top 8 teams proceed to a series of one on one playoff matches to determine the eventual winner. Each playoff match typically consists of three different games, with the best performing agent across all games winning the match.
More details on the competition specifics, including how the playoff rounds are structured, how performance on each game is measured, and other rule technicalities, can be found here \cite{genesereth05,genesereth13}.


\subsection{GVGAI Competition}

The GVGAI competition is another general game playing competition, organised by the Game Intelligence Group of the University of Essex in 2014 \cite{gvgai2014}, although many of the ideas and framework behind the competition were first published the previous year \cite{gvgpdagstuhl}. While this competition was originally only concerned with game playing, additional tracks that instead focus on rule and level generation have also been run. Games used in this competition are defined using the Video Game Description Language (VGDL), an alternative to GDL that allows for the creation of 2D arcade-style video games. This language has been significantly extended over the years to allow for new types of video games, such as the addition of a rudimentary physics engine in 2017 \cite{gvgaiphysics}. Several new games are often added each year to the official GVGAI game collection that is included with the available system code,\footnote{GVGAI Software: http://www.gvgai.net/software.php} but many additional games have been created by other research groups to investigate certain specific hypotheses \cite{gvgaideceptive}.

VGDL is designed primarily for describing real-time games that can contain controllable avatars, stochastic effects and hidden information \cite{vgdl}.
While GDL defines games using logical rules, VGDL describes a game based on the entities and interactions that can occur within it \cite{gvgaioverview}. The dynamics of each component's movement, behaviour and abilities is based on its defined type, which are programmed separately into the GVGAI framework. If a new game requires a component that functions differently to those previously defined, then this must first be added into the system itself. While this approach abstracts out some of the core logic and reasoning from each game's description, the result is a language that is much more flexible and which can ultimately produce a greater set of possible games. Game descriptions defined in VGDL are also typically much shorter and more human-understandable than those in GDL. Each game is also accompanied by one or more level descriptions, that defines how the specified VGDL components should be initially arranged.

The GVGAI competition currently has five distinct tracks, three of which focus on game playing and the other two on content generation. For game playing, there is the single-player planning track, the two-player planning track, and the single-player learning track. The two-player track involves games that can be played by two separate agents, and can be either cooperative or competitive in nature. In the planning tracks, future game states can be simulated to help an agent plan its next actions. The learning track removes this forward model, requiring agents to determine themselves how certain components behave and function. During the competition, agents do not have access to the VGDL descriptions for each game, and receive information only about the current game state. The agent must respond with a discrete action every 40ms, after which the next game state is computed. In addition to these, tracks focusing on level generation and rule generation are also available. In the level generation track, submitted generators must create enjoyable levels for a provided VGDL game description. In the rule generation track, submitted entries must create the rules for a given game, given the available components and level description.
More information about the GVGAI competition and each of its available tracks can be found here \cite{generalgame2,gvgai2014,gvgaitwoplayer,gvgaitwoplayer2,gvgailevelgen,gvgairulegen}.


\begin{figure*}
\centering
      \begin{subfigure}{0.03\textwidth}
    ~
  \end{subfigure}
    \begin{subfigure}{0.47\textwidth}
    \includegraphics[width=\textwidth]{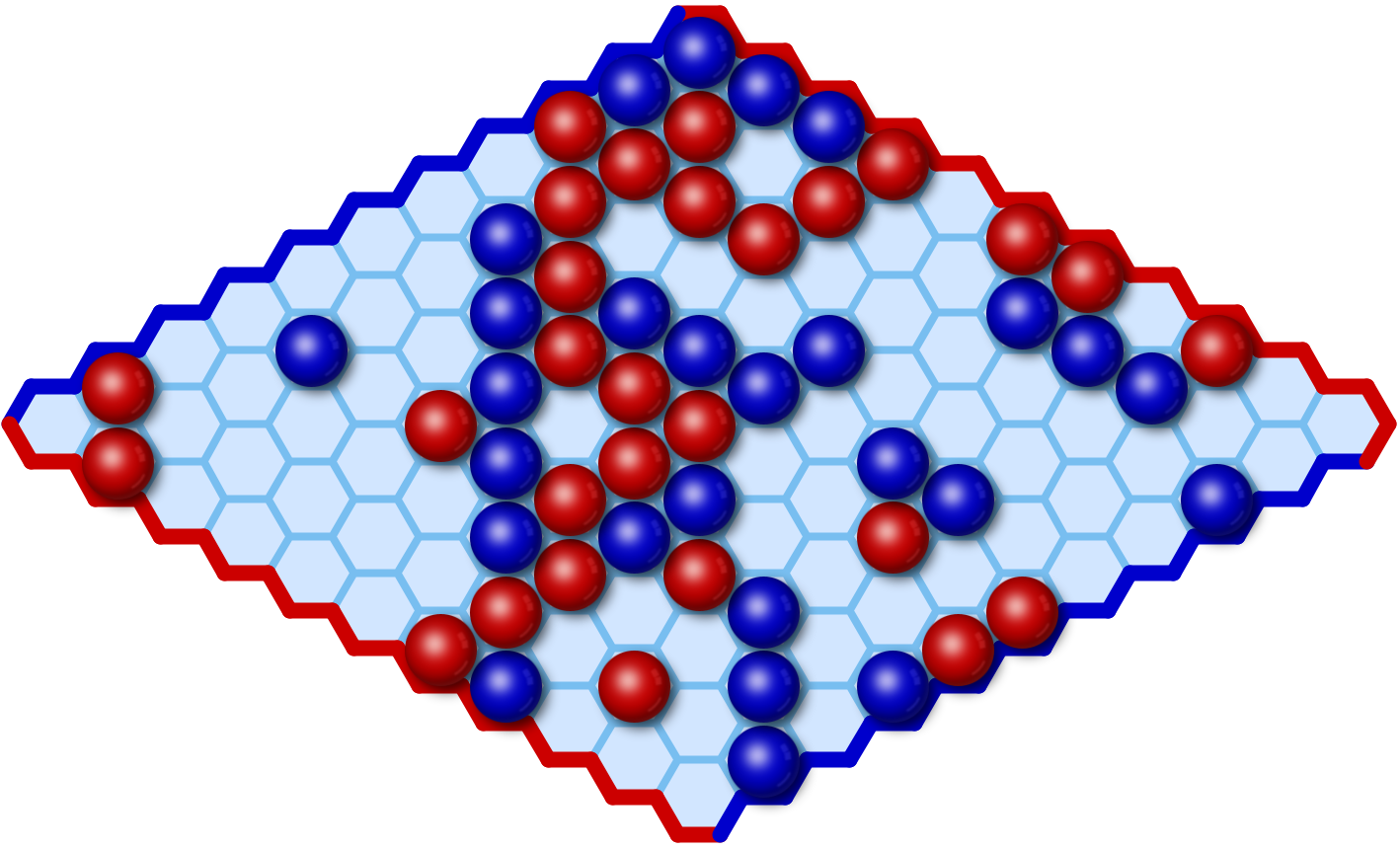}
    \label{fig:1}
  \end{subfigure}
      \begin{subfigure}{0.02\textwidth}
    ~
  \end{subfigure}
  \begin{subfigure}{0.45\textwidth}

\footnotesize
\lstset{basicstyle=\ttfamily}
\begin{lstlisting}
(game "Hex"  
  (mode 2 (addToEmpty))  
  (equipment { 
    (HexBoard 11) 
    (ball Each)
    (region P1 (edge NE)) (region P1 (edge SW))
    (region P2 (edge NW)) (region P2 (edge SE))
  }
  )  
  (rules 
    (play (to (empty)))
    (end 
       (connect (mover))  
       (result (mover) Win) 
    ) 
  )
)
\end{lstlisting}
  \end{subfigure}
  \caption{A completed game of Hex, played out on the Ludii system, along with the its ludeme-based game description.}
  \label{Fig:Gomoku}
\end{figure*}


\section{The Ludii System}

Games can often be broken down into conceptual units of game-related information, known as ludemes \cite{parlett16}. Describing games in this manner is referred to as the ludemic representation. An important benefit of this representation is that it encapsulates key game concepts and gives them meaningful labels, which allows them to be expressed in a compact and human-understandable manner. 
Ludii is a new and complete general game system \cite{browne14}, that can model games as a collection of ludemes using a class grammar approach \cite{BrowneB16}.
This approach allows the game description language to be directly derived from the library class hierarchy, giving a guaranteed 1:1 mapping between the source code and the grammar. Ludii effectively makes its programming language (Java) the game description language, and as a result, it can theoretically support any rule, equipment or behaviour that can be programmed in Java. The exact details of this implementation are hidden from the user, who only deals with the simplified grammar which summarises the code to be called. Games defined using this language can be described or modified in very few lines of code, providing many possibilities for procedural game generation algorithms. As an example, Figure 1 shows the Ludii visuals and ludemic description for the board game Hex, with additional examples available here \cite{Piette19,cognew1,cognew2}


The number and variety of games that can and will be implemented within Ludii also goes far beyond those presented by any previous general game system. 
Ludii can successfully model the full range of GDL games, as well as stacking games, boardless games and even games with hidden information, along with many other possibilities. This, coupled with the high efficiency achieved by Ludii compared to other general systems \cite{Piette19}, makes it one of the largest and most research friendly collections of benchmark games for agents.

In the following sections we propose several future competitions and tracks that we intend to organise and run using the Ludii system. Many of these competition ideas could also be open for humans participants as well as AI. While we intend to run some human exclusive competitions, for the sake of brevity in this paper, we concentrate primarily on competitions that involve the submission of an AI-based agent or content generator.

\section{Agent-based Competitions}
This section describes competitions that focus on playing games. Submitted entries to these competitions must be able to play any game defined within the Ludii system without any external assistance. Rather than describing each of the separate tracks that could be run, we instead describe the different competition requirements and frameworks that may be possible. We expect to run a large number of tracks, each of which constitutes a unique combination of these properties. For example, one track might focus on a particular type of game, where agents are given certain information, and where the competition is structured in a certain format. Another track might change one or many of these competition aspects, giving an entirely new challenge.
It is also possible for a competition track to feature several options for a single property, or to randomly select the properties for each individual match.

\subsection{Game Source}
A large number of Ludii games, defined using Ludii game description language, will be publicly available on the official Ludii repository. The games used in any competition may be sourced from this or may be created especially for it.

\subsubsection{Known Games}
The evaluation process will be conducted using a set of games from the official Ludii repository. Participants will know before the competition begins which games will be used during it.

\subsubsection{Unknown Games}
Same as above, but competition participants do not know the exact games that will be used from the Ludii repository.

\subsubsection{New Games}
Entirely new games or game variants are used to evaluate agents. Participants will not have access to these games before the competition.

\subsection{Types of Games}
Games can often fit into different families or categories, and agents may be required to play games from some of these.

\subsubsection{No holds barred}
Any type of strategic game is possible.

\subsubsection{Specific Games}
All games will belong to one, or maybe several, game categories. The exact game classification scheme is yet to be decided, but likely categories could include:
\begin{itemize}
  \item Deterministic / Stochastic Games
  \item Complete / Hidden Information Games
  \item Board / Card / Dice / Tile Games
  \item Boardless Games (e.g. Dominoes)
  \item Stacking Games (e.g. Lasca)
  \item Simultaneous Move Games (e.g. Chinook)
  \item Graph Games (e.g. Dots and Boxes)
  \item Team Games (discussed further in the next section)
\end{itemize}

\subsection{Number of Agents}
Games can also vary in terms of the number and dynamics of the players involved.

\subsubsection{Multi-player competitive games}
The competition uses games that involve two or more agents competing against each other. The goal is to win more games than your opponent(s), or to achieve the highest average ranking across multiple games. This can also include team games that involve several groups of agents competing against each other.
\subsubsection{Multi-player cooperative games}
The competition uses games that involve two or more agents cooperating with each other to achieve some joint objective. Agents may also have their own individual goals that they are also trying to complete. Points are awarded to each agent based on the number of completed objectives, over several different games and team combinations.
\subsubsection{Single-player puzzles}
The competition uses games / puzzles that involve only one player. The goal is either to be the first agent to solve the puzzle, or to find the shortest possible solution in a given time (if the puzzle has multiple solutions).

\subsection{Information Provided}
Agents may be provided with certain information about the games they are required to play.

\subsubsection{Forward model}
Agents have access to the forward model for each game. This forward model can be used to evaluate future states. Random game elements within these forward models can be simulated automatically or specified by the agent to achieve a desired outcome.

\subsubsection{Game Description}
Agents have access to the Ludii game descriptions. Agents must infer each game's moves, rules and victory conditions from this.

\subsubsection{Going in Blind}
Agents do not have access to either the game description or forward model, and must infer them as they play.

\subsection{Learning Time}
While all public games on the official Ludii repository will be available to all participants, if the games used to evaluate the agents are not known or available beforehand, then a set amount of time may be allocated for the agents to learn about the games they will be required to play.
This is independent of whether the agents have access to the game description, forward model, or neither.

\subsubsection{No Learning Time}
Agents do not get any time to train on the games before the evaluation phase.

\subsubsection{Learning Time for the same games}
Agents have access to the evaluation games for a short period of time (training phase) before they are evaluated. The exact length of time that agents will have access to these games can vary considerably, from a few minutes, to several hours, or even a couple of days.

\subsubsection{Learning Time for different games}
Agents will be provided with a training phase before the evaluation phase, but the games available in each phase will be different in some way. Games provided during the training phase will likely be minor variants of those used in the evaluation phase, (e.g. differed board size/shape, different starting positions, etc.). Agents will need to use transfer learning capabilities to generalise the information obtained from the training phase into the evaluation phase.

\subsection{Competition format}
The first two competition formats we propose here are short-running and intended for one off events held over a few hours or days. The second two formats are long-running and are expected to take place over many weeks or months. We intend to run both annual competitions, and continuous rankings that are updated all year round. Note that for cooperative games or multi-player games with more than two agents, additional scoring considerations will likely need to be made.

\subsubsection{Elimination}
Agents play against each other in a series of knockout style rounds. Each round of the competition selects a new set of games that all remaining agents are then evaluated using. The best performing agents from each round then proceed to the next, until only one agent remains. If the initial number of agents is very high, then a benchmark set of games may be used to select a smaller number of best performing agents that then proceed to the main tournament rounds.
		
\subsubsection{Round-Robin}
Each agent plays against all other agents on a single set of games. Points are awarded for each match-up based on the ``Three points for a win'' system (win = 3 points, draw = 1 point, loss = 0 points). The final point ranking may be used to calculate the winner of the competition, or to determine which agents qualify for a separate elimination format tournament.

\subsubsection{Leaderboard}
Agents can be submitted to a centralised server that keeps a continuously updated ranking of each agent's average performance across all games in the public Ludii repository. Individual performances are recorded for each game, as well as for each game category and across the entire game set. New agents or versions can be submitted at any time, to compare their performance against other existing agents. Humans are also able to be ranked on this leaderboard alongside these agents, as a benchmark measure of human vs AI ability. 
Performance values for each agent will likely be based on their percentage number of wins. Several rating systems for this are possible, such as Glicko-2 which is used by the GVGAI two-player track or AGON which is used for GGP.

\subsubsection{League}
Agents can be entered to compete in a league that runs over a certain period of time, likely a few months. This league operates similar to those of most sports, where all entrants play against each other over a series of matches. This would be conducted in a manner very similar to that of the described Round-Robin format, but where a greater number of games are played less often and over a longer period of time. Development teams might also be able to adjust their agent between games, to fix issues or tailor strategies towards an upcoming opponent. New games are expected to be played each week and will be live streamed online for people to watch. Some of the best performing human players may also be invited to take part in these competitions, to provide a comparison with human capabilities.

\section{PCG-based Competitions}
This section describes competitions that focus on generating games, or specific aspects of games. Similar to the previous section, we describe several different options for certain competition properties. Several adjustable properties of PCG-based competitions, specifically ``Types of Games'' (under which the intended ``Number of Agents'' could also be included), ``Information Provided'' and ``Competition Format'', include the same possibilities as agent-based competitions, and are therefore not repeated.

\subsection{Generatable Content}
What content the generator is expected to create.
Many different game-related aspects can be designed automatically by content generators, spanning from a complete game, including rules, board, pieces, etc., to certain game elements, and even material for indirectly enhancing the game experience. 

\subsubsection{Games}
Generators must create a complete and valid game. The generated game should be described using the Ludii game description language, and must compile successfully on the Ludii system.
\subsubsection{Puzzles}
Generators are given the Ludii description for a puzzle and must generate an initial state. In other words, the generator must create one or more puzzles for a given set of rules. This idea of defining the initial setup for a set of game rules could also apply to multi-player games, but is more applicable to single-player games and puzzles.
\subsubsection{Rules}
Generators are given the equipment that is available for a game (board, pieces, dice, cards, etc.) and must generate a viable ruleset, including game setup, turn process and victory conditions. Generators may be required to use all of the equipment available, or can ignore some if they wish.
\subsubsection{Instructions / Tutorials}
Generators are given the Ludii descriptions for a set of games, and must translate each of these into a plaintext English description. This may be extended to other languages as well, if there is sufficient demand. Similar work on this has been done for VGDL descriptions \cite{tutorialgeneration}, but to our knowledge no competition around this has yet been run. This can also be extended to the generation of full tutorials, with accompanying pictures, videos or game scenarios that demonstrate key rules.

\subsection{Generation Requirements}

\subsubsection{Free Form}
Generators have no, or very few, requirements on what can be generated.

\subsubsection{Specified ludemes}
Generated games or rules must include a certain set of ludemes. Predefined groupings of ludemes (ludemeplexes), that define aspects such as the game board, specific rules, victory conditions, etc., may also be specified as a generation requirement.

\subsubsection{Partial Evidence}
This requirement is similar to that of specifying ludemes, but is based on actual archaeological and anthropological evidence. Generators will be provided with a description of the equipment and rule information that is known for a game, along with the historical and cultural background surrounding it, and must generate a possible game using this information. 

\subsubsection{Deceptive Games}
The term deceptive games has been coined before in relation to several types of games, and can be used to refer to any games that exploit biases and weaknesses within current agent techniques \cite{deceptive1,deceptive2}. Several previous games, such as Arimaa, have already been designed with the goal of being comparatively difficult for AI compared to humans. We propose a competition centered around this, where generators try to design games that current agents perform poorly on, relative to the ability of human players (and vice versa). If successful, such games may help us to better understand the limitations of current AI techniques for general game playing.

\subsection{Evaluation Procedure}
How the generated content should be compared and ranked.

\subsubsection{Judging Panel}
Generated content is evaluated by a panel of human judges. This can be accompanied by computational testing to remove impossible or infeasible possibilities prior to human evaluation.
Each judge will be presented with all or some of the generated content, and will be asked to rank them based on some desired metrics. This could include overall enjoyment, novelty, difficulty, complexity, historical accuracy, etc., depending on the competition specifications.

\subsubsection{Agent Evaluation}
Generated content is evaluated by a collection of agents. This is unlikely to be effective on its own, but can be combined with human evaluation to provide a more detailed evaluation profile. Agents and other AI techniques can also evaluate games in terms of some defined quality metrics, such as fairness, game length, strategic depth, rule complexity, drawishness, etc.

\subsubsection{Public Vote}
Generated content is rated by members of the general public. The exact format for this evaluation, online poll, live voting, user survey, etc., is still to be decided. Like other evaluation procedures, this can be combined with agent-based evaluation and expert judges.



\section{Other Competitions}
This section describes competitions that focus on topics related to both or neither of the previous two sections.

\subsection{Turing Test}
Turing test tracks could be run for both the Agent and PCG based competitions. For the Agent-based competitions, agents would need to play in a human-like and believable manner. Human judges would then play a number of games against each opponent, and attempt to determine if the adversary is an agent or human. The same principle can also apply for PCG-based competitions, where human judges would guess if some content was created by a human or generator. A competition of this nature was held previously as part of the Mario AI competition \cite{marioaicompetition} and some preliminary research has been done for GVGAI games \cite{gvgaituring}.

\subsection{Generate and Play}
Competitions that combine Agents and PCG could be also be held. Each competitor (human or AI) could provide or generate a Ludii game, which are then combined to make up the set of agent evaluation games. Generators or human designers would need to create games that would play to the strengths of their own agent, but which other agents would likely perform worse on.

\subsection{Stab in the Dark}
A designer presents a game that human participants must then play without seeing the rules. These participants play against an AI to learn the game, and try to beat it in the smallest amount of time or number of moves. Participants may be told some small amount of information about the game, such as the possible movements of a selected piece, or may be told nothing at all.
The designer who produces the simplest game, which uses the fewest ludemes or has the smallest state/action space, that confuses the most people for the longest time wins.

\subsection{Needle in a Haystack}
Participants submit a game evaluation system that is given 1,000 to 1,000,000 randomly generated games or rule sets to test. These evaluation systems are then given a certain period of time to rank these games and select those which it believes are best, according to their own criteria. Expert judges or members of the general public will then play and rate the best games selected by each agent, to decide the  winner.



\subsection{QR-code design}
File sizes for games defined in Ludii are so small, that each game description can likely fit onto a single QR code. Inspired by similar work on designing custom QR codes artwork,\footnote{https://research.swtch.com/qr/draw} we propose a competition around designing the most visually appealing QR code for a playable Ludii game (e.g. Chess with a picture of a piece, or Hnefatafl with an image of a viking).



\section{Conclusions and Challenges}
In this paper we have described the current area of General Game Playing, and presented Ludii as an alternative system with many new ideas for game AI research and future competitions.
Ludii provides many interesting problems and challenges for game researchers.
Ludii offers a sophisticated system and language, as well as a comprehensive database of traditional strategy games. Ludii is also efficient, understandable, modifiable, and convenient to work with. 
It provides an ideal platform for playing, generating and testing games, that is fully compatible with many AI techniques.
Apart from a purely academic platform, Ludii has the potential to be an important game design tool that can assist human game designers with the creation, analysis and play testing of their games








With such a large range of potential competitions it is important to focus on those that will have the largest development and research base, and which can also provide the most interesting results for AI research. It will therefore be impossible to cover every possible competition possibility, but we hope to identify the most popular and promising tracks over the next few months. Other challenges include internally developing and testing the large number of game implementation needed. While it will be possible for anyone to create their own games using Ludii for testing purposes, the official Ludii game repository must be subject to a rigorous quality standard. Running such a large number of competitions will require substantial hardware and organising time, likely resulting in new competitions being slowly rolled out over the next few years rather than all at once.

Future work over the next few years will primarily involve developing more games and additional functionality for the Ludii system, as well as organising and running a large number of the competitions proposed in this paper.
As well as traditional strategy games, we hope to develop many other types of games for Ludii. This may include many modern board game ideas, simple video games and serious games.
Looking even further into the future, we anticipate branching out into other non-game domains and applying Ludii to problems such as protein folding, physics simulations or predicting chemical reactions.




\section*{Acknowledgment.}

This research is part of the European Research Council-funded Digital Ludeme Project (ERC Consolidator Grant \#771292) run by Cameron Browne at Maastricht University's Department of Data Science and Knowledge Engineering. 

\bibliographystyle{IEEEtran}
\bibliography{References}

\end{document}